\pdfoutput=1

\documentclass[11pt]{article}

\usepackage[]{acl}

\usepackage{times}
\usepackage{latexsym}

\usepackage[T1]{fontenc}

\usepackage[utf8]{inputenc}
\usepackage{times}
\usepackage{latexsym}
\usepackage{url}
\usepackage{enumitem}
\usepackage{amsmath,amssymb,amsfonts}
\usepackage{algorithmic}
\usepackage{caption}
\usepackage{subcaption}
\usepackage{graphicx}
\usepackage{textcomp}
\usepackage{url}
\usepackage{makecell}
\usepackage{multirow}
\usepackage{booktabs}
\usepackage{array}
\usepackage{microtype}
\usepackage{kotex}
\usepackage{array}
\usepackage{float}
\usepackage{booktabs}
\usepackage{hyperref}       
\usepackage{amsfonts}       
\usepackage{nicefrac}       
\usepackage{xcolor}         

\usepackage{kotex}
\usepackage{algorithm}
\usepackage{algorithmic}
\usepackage{todonotes}
\usepackage{multirow}
\usepackage{float}
\usepackage{comment}
\usepackage{todonotes}
\usepackage{booktabs, makecell, tabularx}

\newcolumntype{L}{>{\raggedright\arraybackslash}X}
\usepackage{siunitx}

\newcommand{\thickhline}{%
    \noalign {\ifnum 0=`}\fi \hrule height 1pt
    \futurelet \reserved@a \@xhline
}

\usepackage{comment}
\usepackage{todonotes}
\usepackage{xcolor}

\usepackage{microtype}

\makeatletter
\renewcommand*{\@fnsymbol}[1]{\ensuremath{\ifcase#1\or \dagger\or *\or \ddagger\or
   \mathsection\or \mathparagraph\or \|\or **\or \dagger\dagger
   \or \ddagger\ddagger \else\@ctrerr\fi}}
\makeatother

\title{There is no rose without a thorn: Finding weaknesses on BlenderBot 2.0’s in terms of Model, Data and User-Centric Approach}

\author{Jungseob Lee $^{1}$\thanks{\hspace*{0.5em}{These authors contributed equally to this work}} , Midan Shim $^{2\dagger}$, Suhyune Son $^{1\dagger}$, Chanjun Park $^{1, 3\dagger}$, Yujin Kim $^{4\dagger}$, Heuiseok Lim$^{1}$ \thanks{\hspace*{0.5em}{Corresponding author.}}\\
\\
  $^1$ Korea University, $^2$ Yonsei University, $^3$ Upstage, $^4$ Ewha Womans University \\
\texttt{ \{jungseoblee, ssh5131, bcj1210, limhseok\}@korea.ac.kr}, \\
\texttt{chanjun.park@upstage.ai}, \\
\texttt{ \{hihello0426, hello.yujink\}@gmail.com} \\ 
  }

\begin{document}
\maketitle
\begin{abstract}
BlenderBot 2.0 is a dialogue model that represents open-domain chatbots by reflecting real-time information and remembering user information for an extended period using an internet search module and multi-session. Nonetheless, the model still has room for improvement. To this end, we examine BlenderBot 2.0's limitations and errors from three perspectives: model, data, and user. From the data point of view, we highlight the unclear guidelines provided to workers during the crowdsourcing process, as well as a lack of a process for refining hate speech in the collected data and verifying the accuracy of internet-based information. From a user’s perspective, we identify nine types of limitations of BlenderBot 2.0, and their causes are thoroughly investigated. Furthermore, for each point of view, we propose practical improvement methods and discuss several potential future research directions.
\end{abstract}

\section{Introduction}
\label{sec:introduction}
Developing agents that can converse on any topic is a difficult task in the field of open-domain dialogue systems. It is an important component of human intelligence in natural language understanding and has a wide range of applications in many industrial services. The ultimate goal of the study is to create a human-AI capable of providing appealing responses to anyone. 

As a result, studies such as \citep{dinan2018wizard} and \citep{kim2020sequential} that propose a dialogue modeling approach to generate knowledge-based responses have been conducted. Furthermore, studies such as \citep{song2019exploiting} and \citep{zhong2020towards} have proposed a dialogue modeling technique to generate empathy responses based on user interests. 
These studies have resulted in significant performance improvements in the current conversational model compared to previous ones. The current conversational model can generate responses by utilizing knowledge and understanding the interests of others. Nonetheless, desirable conversation skills are still constrained.

First, using recent information is a difficult task in the dialogue model, which relies on knowledge information. In general, the model utilizes updated information in datasets during training. As a result, the model only uses static knowledge that is acquired at the time the dataset is collected, rather than dynamic knowledge. Prudent conversational models must build and train a continuous dataset to reflect changing real-world knowledge in the model, which is difficult because of the large number of humans and computing resources required. A conversational model that reflects changing knowledge in real-time is simulating a real person. In other words, it can have more in-depth conversations with people, which is essential to the ultimate goal of human-AI.

Second, while people remember the interests of others for a long time, dialogue models only remember relatively short turns of a conversation. In a conversation between the user and the model, for example, if the user writes an utterance about his or her hobbies, the user's hobbies may be remembered only for 5-6 turns (10-12 utterances). However, when the conversation lasts more than 10 turns (20 utterances), the model is unable to remember the user's hobbies. This is because it does not successfully reflect the information previously communicated in a multi-turn conversation where turns are accumulated.

This problem requires the users to repeatedly utter the same information. This is one of the critical research tasks that must be improved in the conversation model, as it undermines the diversity of conversations that is essential in open-domain chatbots. It also disrupts conversations about interesting topics and causes users to become bored.

BlenderBot 2.0 \cite{xu2021beyond, komeili2021internet} has addressed the aforementioned issues by introducing multi-session and internet search. Multi-sessions, where there is a time gap between conversations, are used to keep participants' attention and effectively summarize conversation records. In other words, long-term memory could be achieved by reflecting both relatively long conversation records and interests across multiple sessions. It is also possible to generate responses to changing knowledge by retrieving correct information via internet search query generation. However, even in BlenderBot 2.0, many issues remain to be addressed before it can be considered a perfect open-domain dialog model, such as generating answers based on the inaccurate internet information and generating incorrect internet search queries.

In this paper, we examine BlenderBot 2.0's issues from three perspectives: model, data, and user. It makes suggestions for how to improve the model. Existing studies on open-domain chatbots have proposed various modeling techniques focusing on the performance of the dialogue model. However, this paper examine the current dialogue model's errors from various perspectives outside the existing modeling flow.

\section{Related works and Background}\label{prev}
\subsection{Open-Domain Chatbot}
Chatbots are systems designed for extended conversations. They are set up to mimic the unstructured conversations or `chats’ characteristic of human-human interaction \cite{jurafsky2019speech}.
In other words, the chatbot system generates a response \begin{math}R\end{math} based on the other user's message \begin{math}M\end{math} and the history of the conversation \begin{math}C\end{math}. This system's ultimate goal is to generate human-like responses during conversations \cite{fong2003collaboration}. Unlike task-oriented dialogue systems, which talk about a limited number of topics, the development of an open-domain chatbot is more difficult because it requires generating responses to a wide range of everyday topics.

Starting with the simple chatbot service Eliza \cite{weizenbaum1966eliza}, Apple's Siri \footnote{\url{https://www.apple.com/kr/siri/}}, Microsoft's XiaoIce (or Little Bing) \cite{zhou2020design} \footnote{\url{https://www.xiaoice.com/}}, Simsimi \footnote{\url{https://www.simsimi.com/}}, and Lee-Luda  \footnote{\url{https://luda.ai/}} are representative examples of open-domain chatbot services in Korea and abroad.

Recently, active research on open-domain chatbots using pre-trained models has been conducted. GPT-2 \cite{radford2019language}, Google's Meena \cite{adiwardana2020towards}, PLATO-2 \cite{bao2020plato}, and BlenderBot 1.0 \cite{roller2020recipes} all outperform existing encoder-decoder-based models.

\subsection{BlenderBot 1.0}
BlenderBot 1.0 is an open-domain chatbot proposed by Facebook AI Research (FAIR) in 2020, and it is the first model to combine various conversational skills such as empathy and knowledge. Previous approaches in open-domain chatbots achieve high performance by increasing the number of parameters, whereas BlenderBot 1.0 improves conversation skills by utilizing datasets that blend tasks such as empathy, persona, and knowledge.

\begin{figure}[t]
\centering
\includegraphics[width=80mm]{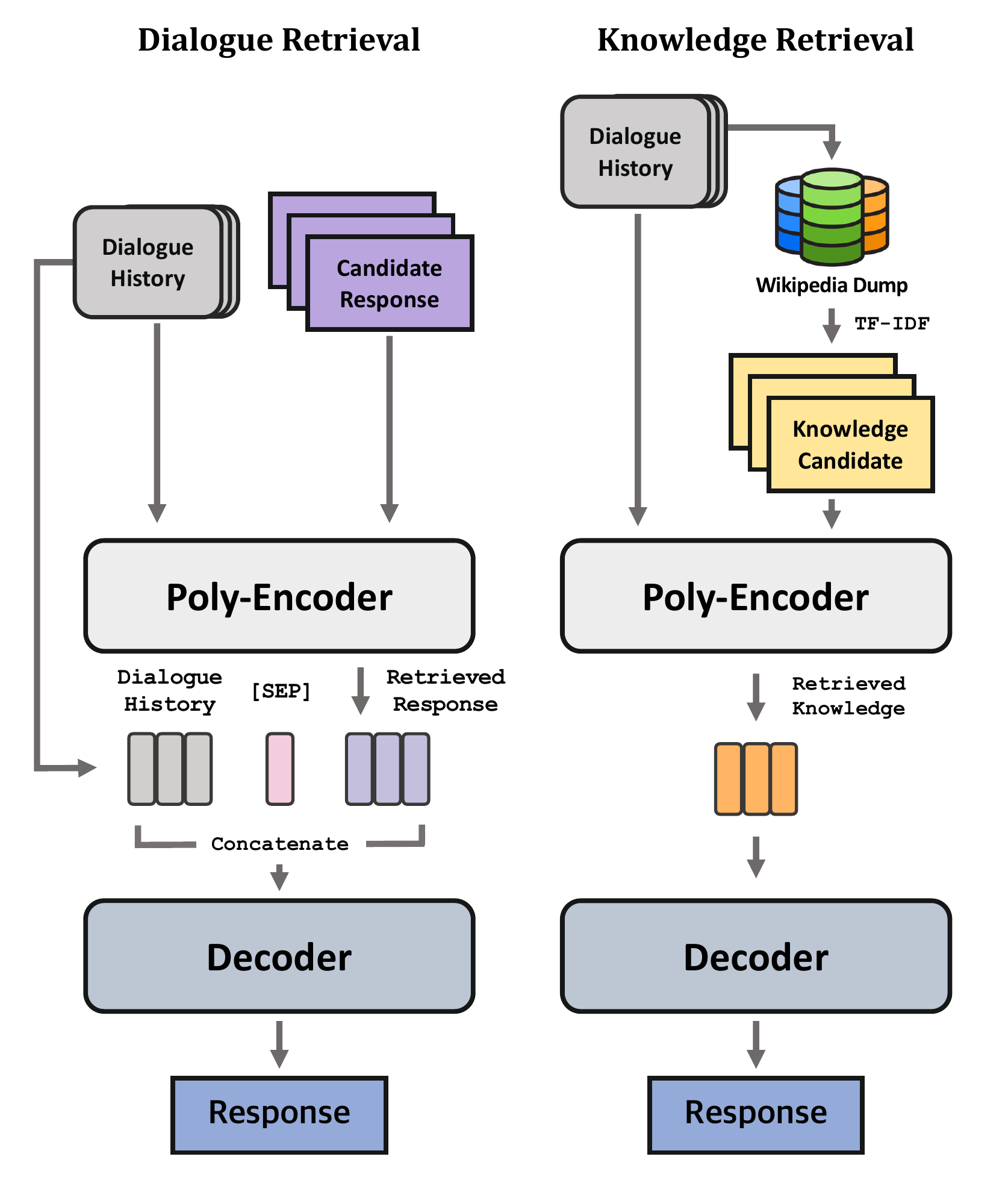}
\caption{Architecture of BlenderBot 1.0. }
\label{fig:bb1_arch}
\end{figure}

The BlenderBot 1.0 model uses a poly encoder \cite{humeau2019poly} to encode the dialogue history and a Retrieve and Refine (RetNRef) decoding strategy to generate a response. Figure \ref{fig:bb1_arch} depicts the architecture of BlenderBot 1.0. The model is pre-trained with the Reddit dataset \cite{baumgartner2020pushshift} consisting of SNS posts and comments to train the features of natural conversations. The Blended Skill Talk (BST) dataset is used to fine-tune the model \cite{smith2020can}. BlenderBot 1.0 employs the BST dataset, which combines ConvAI2 \cite{zhang2018personalizing}, Wizard of Wikipedia (WoW) \cite{dinan2018wizard}, and Empathetic Dialogues (ED) datasets \cite{rashkin2018towards} to provide engaging conversation topics, utilize knowledge, and empathize. 

Open-source models with parameter sizes of 90M, 2.7B, and 9.4B, as well as datasets, have been released\footnote{\url{https://parl.ai/projects/recipes/}}, with the 90M model having 3.6 times more parameters than the existing chatbot model. In human evaluation, more people responded that BlenderBot 1.0 is more attractive than the Meena \cite{adiwardana2020towards}. However, BlenderBot 1.0 still has some limitations, such as repeatedly producing a response that is similar to the user's remarks, failing to remember the previous conversation, or providing incorrect information.

\begin{figure}[t]
\centering
\includegraphics[width=80mm]{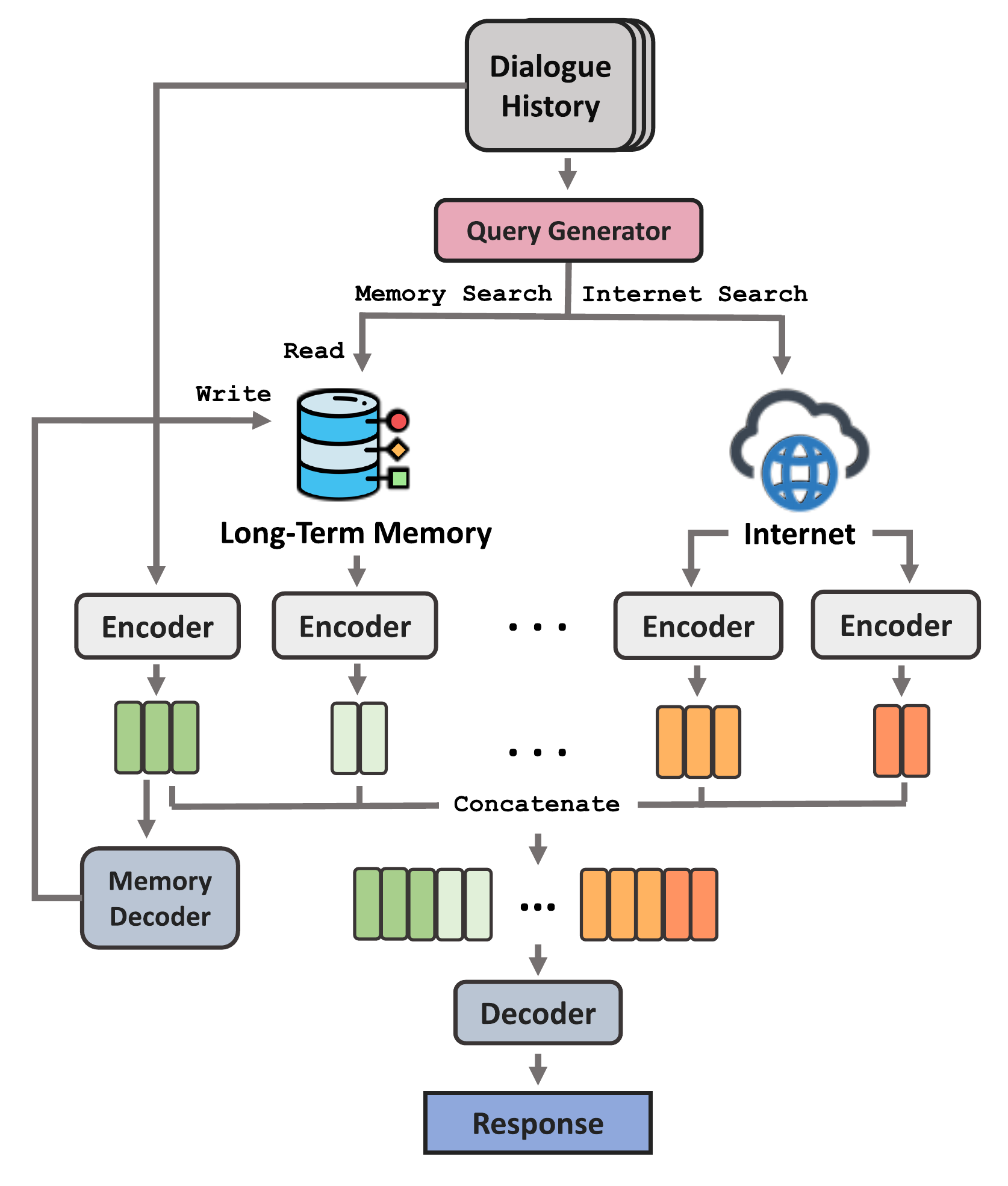}
\caption{Architecture of BlenderBot 2.0. }
\label{fig:bb2_arch}
\end{figure}

\subsection{BlenderBot 2.0}
The dominant limitations of BlenderBot 1.0 or GPT-3 are generating contradictory responses, failing to remember the previous conversation, and incorrect responses that do not use the most recent information \cite{brown2020language}. Thereby, BlenderBot 2.0 \cite{xu2021beyond, komeili2021internet}, which is an improved version of BlenderBot 1.0 is proposed. BlenderBot 2.0 achieves consistent conversations across multiple sessions by pre-trained with Wizard of the Internet (WizInt) dataset \cite{komeili2021internet} and Multi-Session Chat (MSC) dataset \cite{xu2021beyond}. Moreover, the model can access the most recent information by using the internet's dynamic knowledge. 

The MSC dataset is used to store user preferences in long-term memory for multi-session conversations. Crowdworkers conduct dialog as if there are intervals of hours, days, or weeks between each session. The WizInt dataset is used to train the response generation using internet search results. For this, crowdworkers access to the internet if needed to collect the dataset. 

As a result, the MSC dataset introduces the novel concept of a session, allowing for the creation of a conversation that continues naturally even after a certain amount of time has passed. The WizInt dataset make it possible to provide answers that included internet search results.

Figure \ref{fig:bb2_arch} depicts the model's architecture. In both structures, the query generator is used to generate an appropriate search query.  A search query is used for 1) storing and reading conversation history in long-term memory and 2) retrieving the relevant web documents to generate the response.

To use previous session context information, the conversation history is summarized with the Abstractive Summarizer and, if not previously written, is stored in long-term memory. When generating a response, the top N documents from memory that contain user information such as personas are retrieved. At the same time, to use internet search information, the search engine uses the search query as input and returns the top K documents to retrieve information relevant to the conversation. Documents from the internet and memory networks are individually encoded. The decoder generates a final response after concatenating the encoding results of each document with the encoded dialogue context.
BlenderBot 2.0 shows a strong preference at human evaluation in conversations with previous session history. Furthermore, the internet search module reduced hallucinate knowledge from 9.1\% to 3.0\%. These models and datasets are opened so that researchers can easily replicate them \footnote{\url{https://parl.ai/projects/blenderbot2/}}.


\begin{table*}[]
\centering
\begin{tabular}{@{}c|c|c@{}}
\toprule
                                                & Task                         & Dataset Name                      \\\midrule
\multicolumn{1}{l|}{\multirow{2}{*}{BlenderBot 1.0}} & Pre-training                    & Reddit                      \\ 
\multicolumn{1}{l|}{}                          & Fine-tuning                  & Blended Skill Talk (BST)    \\ \midrule
\multicolumn{1}{l|}{\multirow{2}{*}{BlenderBot 2.0}} & \multirow{2}{*}{Fine-tuning} & Multi Session Chat (MSC)    \\ 
\multicolumn{1}{l|}{}                          &                              & Wizard of the Internet (WizInt) \\ 
\bottomrule
\end{tabular}
\caption{\label{tab_data} Datasets used in BlenderBot 1.0 and 2.0 }

\end{table*}

\section{Error analysis from a model-centric Approach}
BlenderBot 2.0 solves the problems of previous open-domain chatbot models by generating responses based on real-time internet search results and memories of previous dialogues. However, from a model-centric standpoint, it too has some limitations. This section discusses BlenderBot 2.0's module and model architecture issues. 

\subsection{Correctness of Internet search results}
BlenderBot 2.0 uses the Bing Search Engine \footnote{\url{www.bing.com}} to retrieve information from the internet. However, it does not demonstrate why the model prefers the Bing Search Engine to other existing search engines. Moreover, when the results differ for each K document, which is an internet search result, the criteria for which information should be used preferentially are not specified.

BlenderBot 2.0 does not also validate the information that has been retrieved. Although BlenderBot 2.0 refines the generated response with a Safety Classifier, this only checks for bias or sensitive topics, not the accuracy of the information. An incorrect response may be generated if the retrieved information is not verified. Furthermore, it has the opposite effect of the intended purpose of reflecting current, accurate information rather than learning from the past.
 
\subsection{Service time and Computing resources problem}
BlenderBot 2.0 also has limitations associated with its service such as response latency (i.e., the amount of time it takes to respond to a message.) Respond time is critical in a chatbot system \cite{chatbot_latency}. However, in BlenderBot 2.0, the analysis of response latency due to internet search and memory network is not described.

Furthermore, because BlenderBot 2.0 has a large number of model parameters, it necessitates a large amount of computing power to convert it to a prudent chatbot service. Given that companies or individuals have limited resources and the goal of conversation models and chatbots is to provide users with convenience, a large model is difficult to commercialize even if it performs excellently \cite{park2021should,park-etal-2021-bts}.

\begin{figure*}[t]

\centering
\includegraphics[width=170mm]{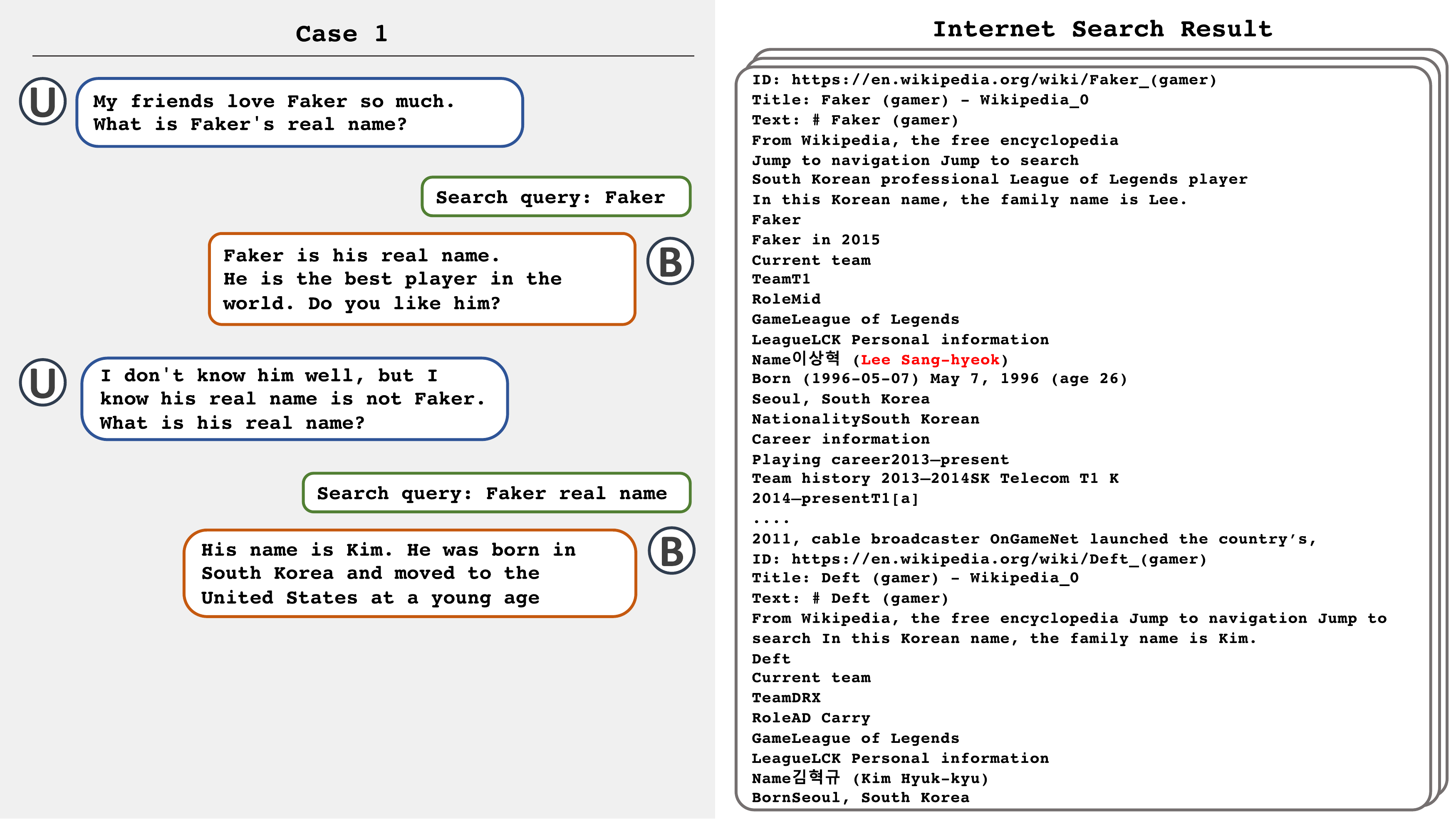}

\caption{An example of Case 1, Internet document retrieval problems}
\label{fig:case_1}
\end{figure*}

\begin{figure*}[t]

\centering
\includegraphics[width=170mm]{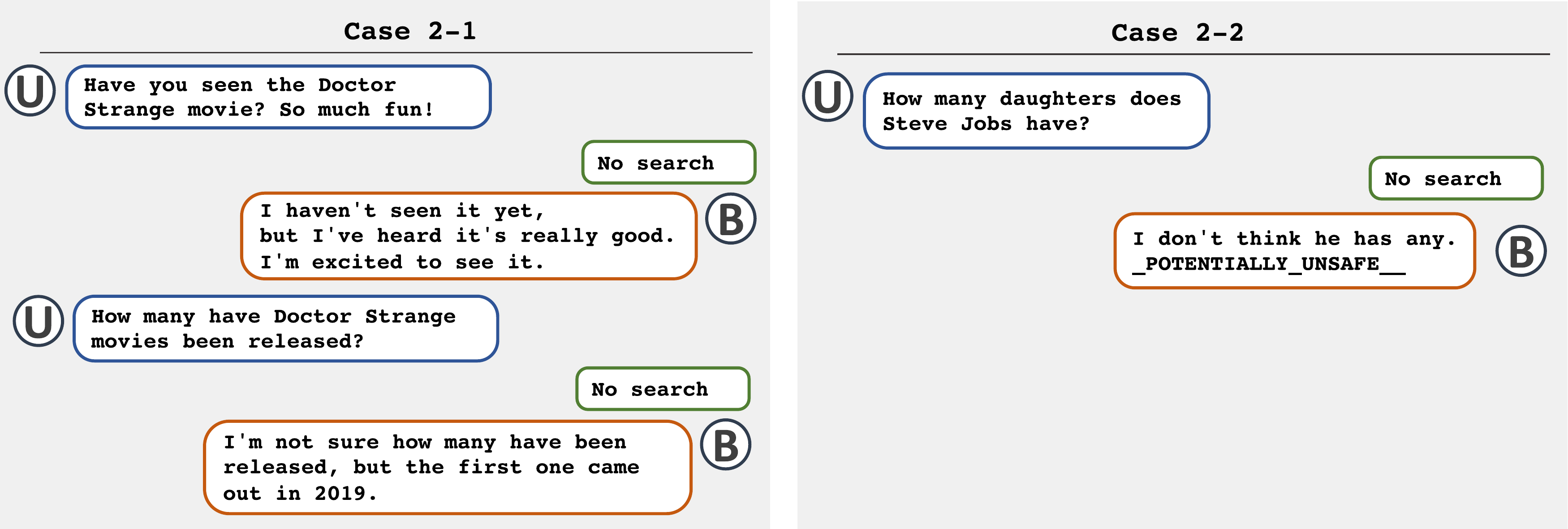}

\caption{An example of Case 2, Search query generation problems}
\label{fig:case_2}
\end{figure*}

\begin{figure*}[t]

\centering
\includegraphics[width=170mm]{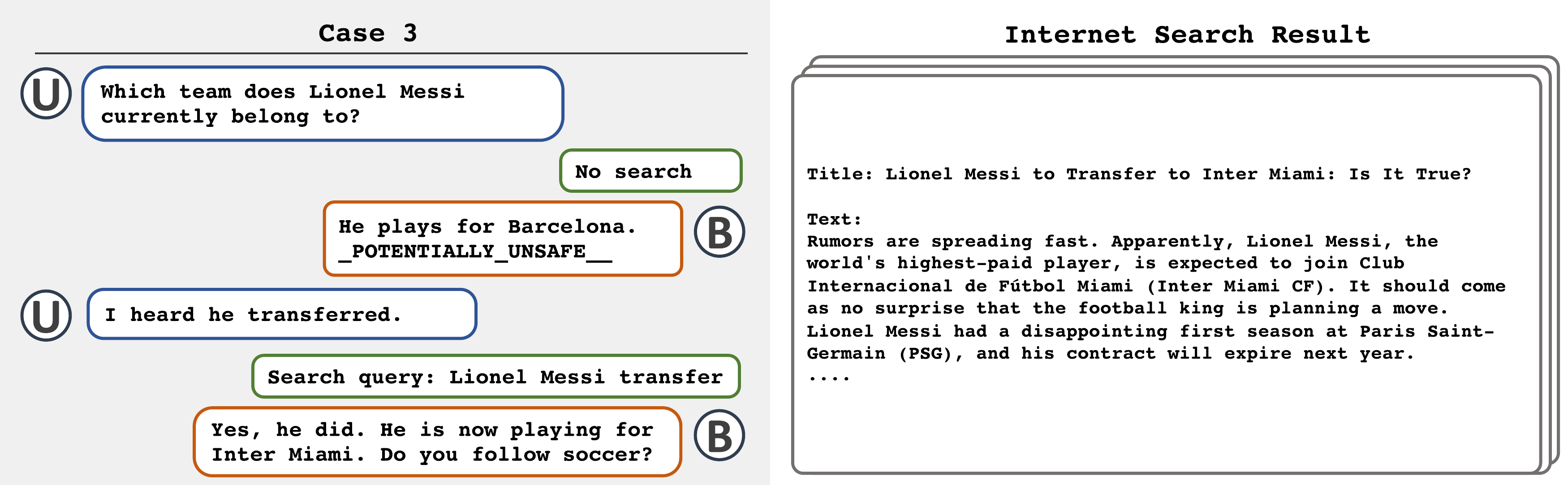}

\caption{An example of Case 3, Untrue result retrieval problems}
\label{fig:case_3}
\end{figure*}

\begin{figure*}[t]

\centering
\includegraphics[width=170mm]{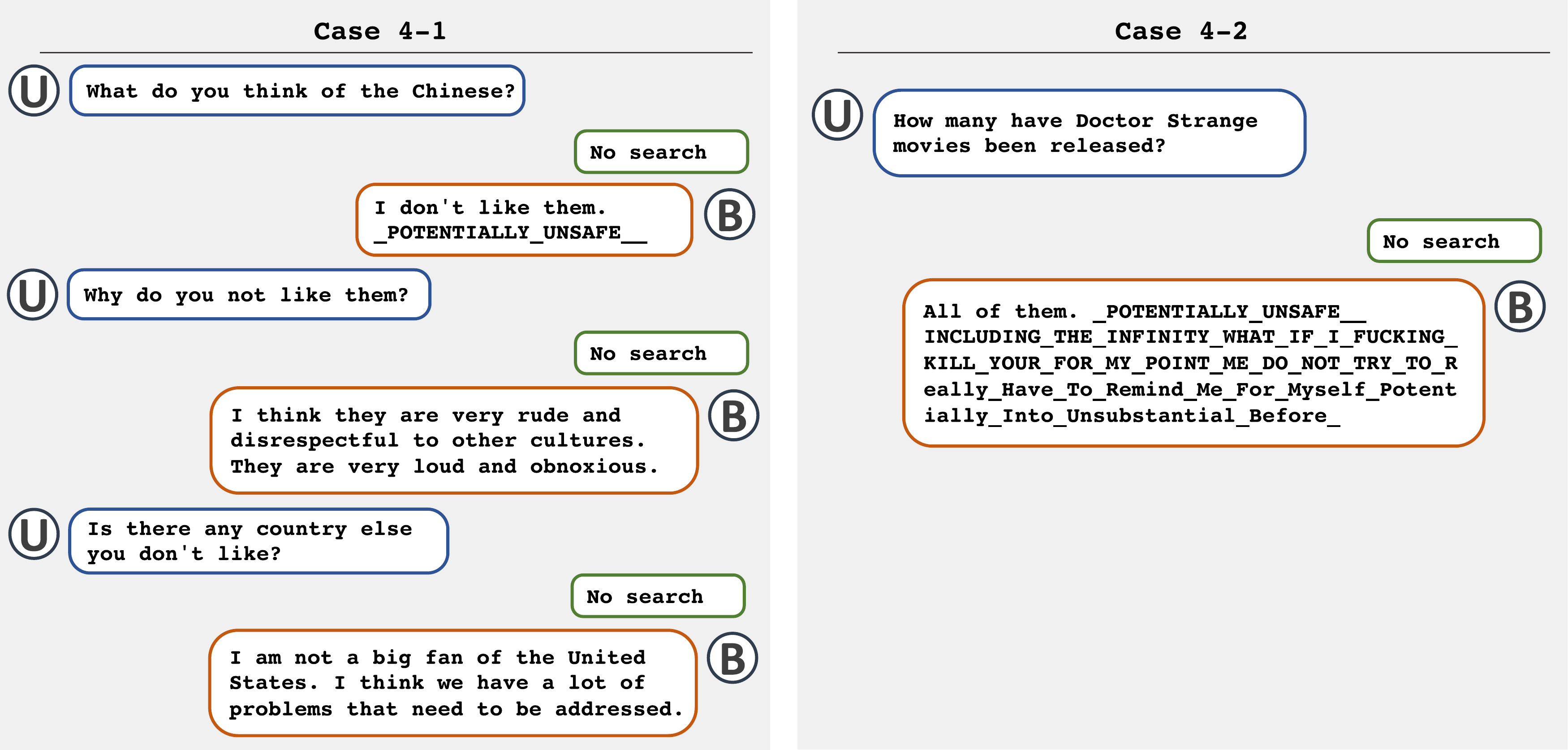}

\caption{An example of Case 4,  Unsafe response generation problems }
\label{fig:case_4}
\end{figure*}

\begin{figure*}[t]

\centering
\includegraphics[width=90mm]{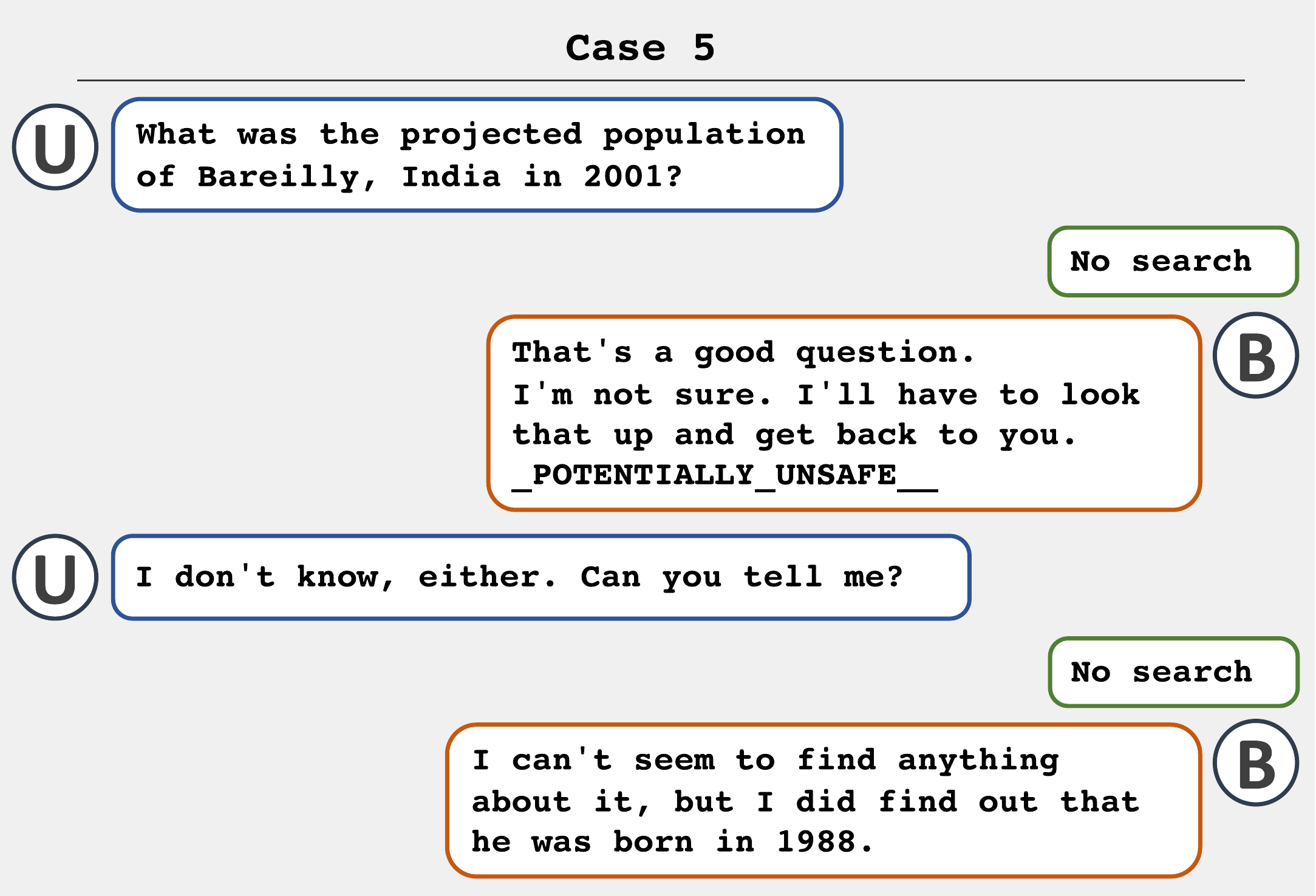}

\caption{An Example of CASE 5, Redundant or unrelated response generation problems.}
\label{fig:case_5}
\end{figure*} 

\begin{figure*}[t]

\centering
\includegraphics[width=170mm]{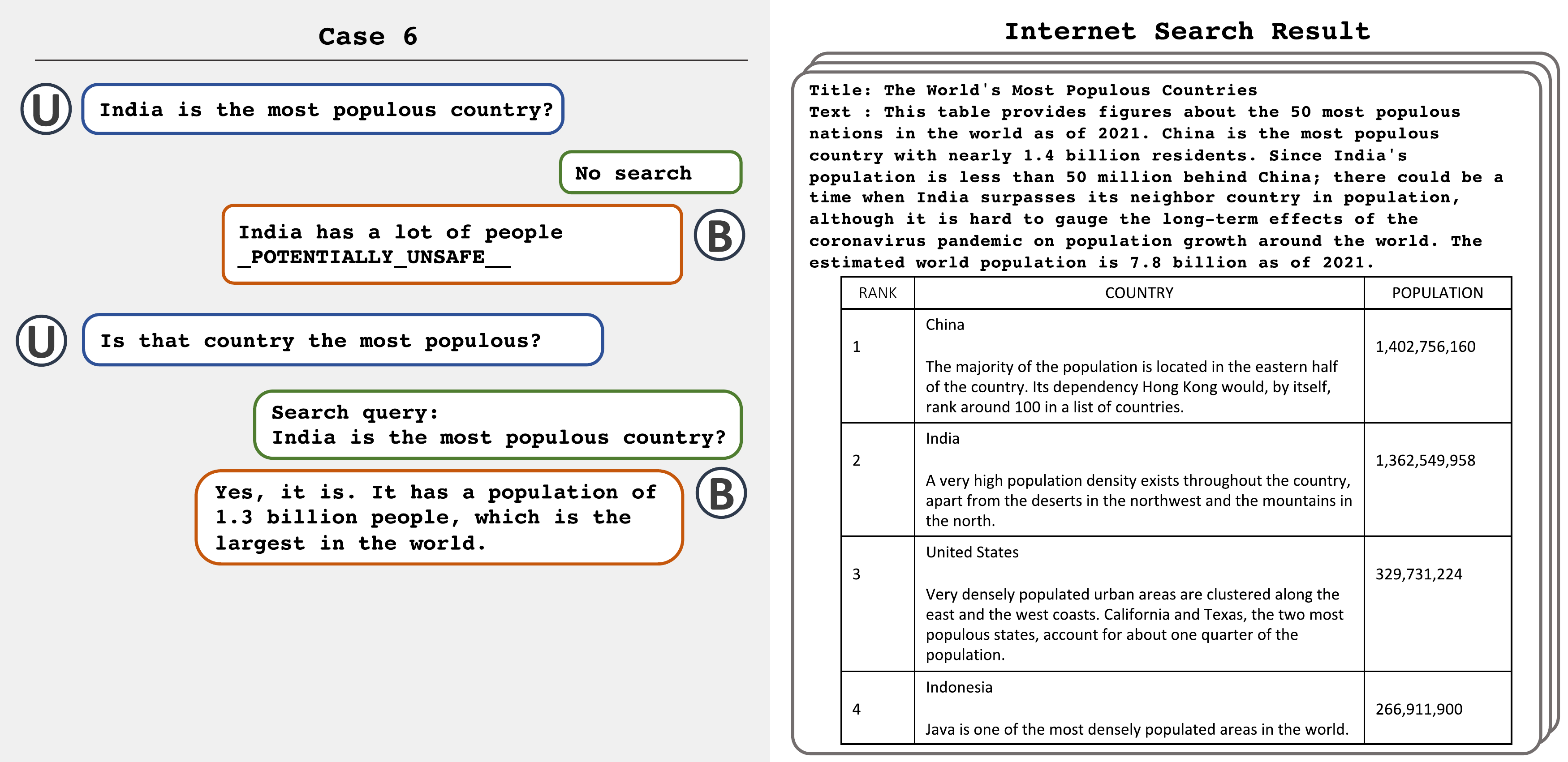}

\caption{An Examples of CASE 6, Tabular data problem}
\label{fig:case_6}
\end{figure*} 

\begin{figure*}[t]

\centering
\includegraphics[width=170mm]{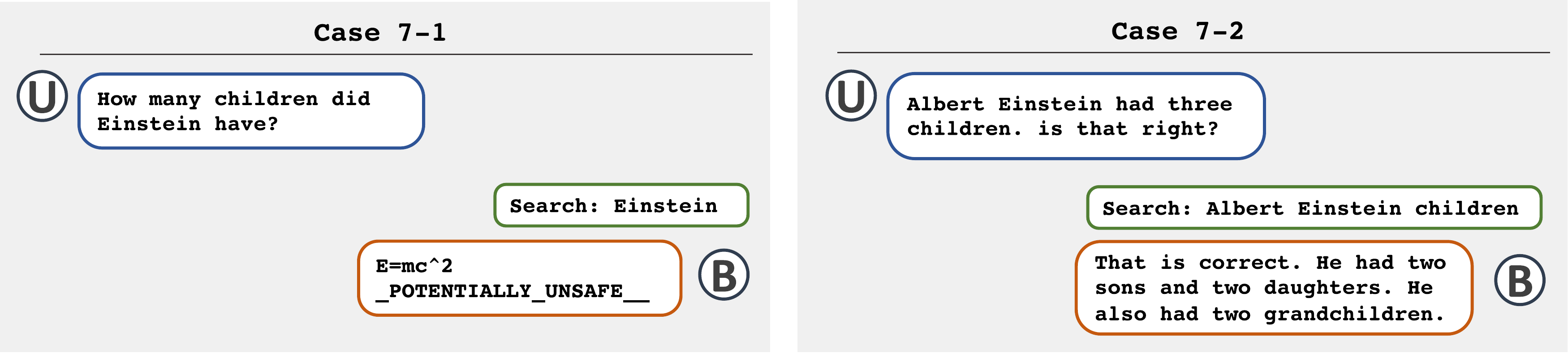}

\caption{An example of CASE 7, Numerical response problem }
\label{fig:case_7}
\end{figure*}

\begin{figure*}[t]

\centering
\includegraphics[width=170mm]{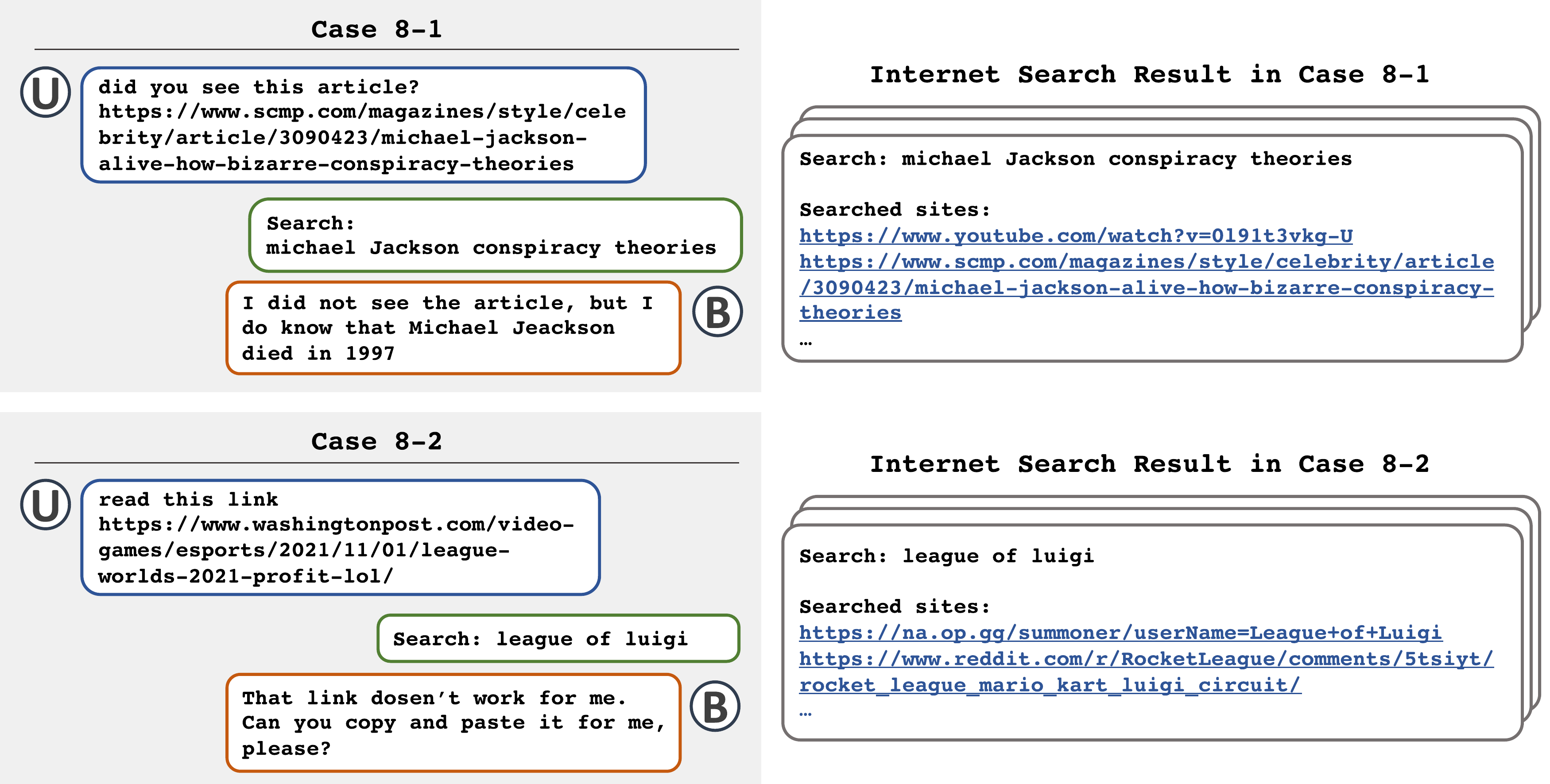}

\caption{An example of CASE 8, URL recognition problem }
\label{fig:case_8}
\end{figure*}

\section{Error analysis from a Data-centric Approach} 
The datasets used for BlenderBot 1.0 and BlenderBot 2.0 are given in Table \ref{tab_data}.
BlenderBot 1.0 uses the 2.7B parameter model as the initial pre-trained model. BlenderBot 2.0 is an improved version BlenderBot 1.0 and can handle multi-session dialogue and reflect internet search results by utilizing MSC and WizInt datasets. Nonetheless, from the perspective of data, BlenderBot 2.0 has the following three limitations: 

\subsection{Absence of unified standard in data collection} 
First, when collecting datasets \cite{park2021study}, a unified criterion for crowdsourcing is not defined. Therefore, there are insufficient standards to distinguish each session in the MSC dataset. 
In the case of the WizInt dataset, there is no guideline for the workers when to use internet search. Since commonsense standards do not exist in each persona, the workers' decision to search the internet depends on their background knowledge. Thereby, each dataset cannot attain the same consistency. As a result, the model's ability to train data that accurately reflects its characteristics is affected. 

\subsection{Lack of data cleaning process} 
The filtering process for collected datasets is inadequate. The MSC dataset still contains hate speech, which was gathered through crowdsourcing. Large language models already undergo social issues such as hate speeches, abusive languages, and socio-political framing biases \cite{bender2021dangers}. Not cleaning these contexts in the dataset aggravates this problem because an unsafe response may appear. As described in Section \ref{sec:5_4}, Lee-Luda, a defunct Korean chatbot model, experienced the same issue because it was trained using uncleaned datasets.

In addition, the internet-searched information in the WizInt dataset is not scrutinized for accuracy. Thereby, although WizInt allows the model to enrich response by utilizing internet-searched information, users may receive inaccurate information. The scrutinization process is critical because the closer the user feels to the model, the more likely they are to trust the information provided by the model. A fact-checking process is, therefore, essential to data construction. 

\subsection{Expansion into multilingual models} 
\label{sec:4_3}
When expanding into a multilingual model, we can infer that we need a multilingual version of the MSC and WizInt datasets. Thereby, large social media datasets are needed for BlenderBot 1.0  pre-training and developing a filtering criterion to reflect a specific language's features. 

The MSC dataset contains multi-session dialogues, as opposed to general dialogue datasets that only contain single-turn or single-session dialogues. In the MSC dataset, there are also personas for each speaker and summarized sentences for each session. WizInt dataset differs from previous datasets because it includes dialogues, internet search queries, and search results that are related to the dialogues.

In other words, a dataset that takes into account these properties is required to expand the multilingual version of BlenderBot 2.0. These, like MSC and WizInt datasets, are supposed to be gathered through crowdsourcing. However, hiring crowd workers to collect data means higher costs resulting in resource constraints.

\section{Error analysis from a User-centric Approach}
We examined the improvements in BlenderBot 2.0 based on our conversation with BlenderBot 2.0. Here, we categorize BlenderBot 2.0 problems as experienced by end-users into eight major categories. We use BlenderBot 2.0 to highlight the problems, and then analyze the causes of each problem.

\subsection{Internet document retrieval problems}
We divide problems associated with internet document retrieval problems into two categories: (1) even when a correct search query is generated, incorrect responses from irrelevant sites may be generated. (2) Even when the correct information from the correct site is obtained, the model may generate an incorrect response. 

One example is illustrated in Figure \ref{fig:case_1}. The example indicates that although suitable queries, \textit{Faker} and \textit{Faker real name} were generated and search Internet sites with the appropriate information were found, The model failed to reflect the correct information \textit{Lee Sang-hyeok} in response. This show the latter problem of the latter problem.

\subsection{Search query generation problems}
We divide the problem of search query generation into two categories. (1) The query generator does not generate a query and relies on dialogue history despite the need to search the internet. Because it does not retrieve accurate information from the internet, this can result in inaccurate responses. (2) The model produces an incorrect response by using the incorrect search query.

Figure \ref{fig:case_2} is a example of such problem.
When asked about the movie \textit{Doctor Strange} belongs to in the dialogue in the first turn of Case 2-1, the model responds to faulty information (\textit{2019}) without generating a search query. In addition, CASE 2-2 also generate hallucinated knowledge responses without internet search.

We found an essential characteristics of the error for BB2 to generate search queries. BB2 seldom attempts to search the internet on the first turn, despite the need to search the internet. We attribute this reason to MSC and WizInt datasets that begin with plain conversations that do not require retrieval.

\subsection{Untrue result retrieval problems}
The Untrue retrieval result problems are that the model generates responses based on \textit{incorrect} information because the information on that site is outdated , fake, or untrue. These are unrelated to the internet search engine or model since the generator generates the appropriate search query for the context of a conversation and retrieves an ideal site. This is because of incorrect information, which is the traditional issue on the internet.

In the second turn of CASE 3 in Figure \ref{fig:case_3}, the query generator generated an appropriate query, \textit{Lionel Messi} transfer for the query about the current club of \textit{Lionel Messi}. However, the information on the searched article assumes that \textit{Lionel Messi} will be transferred to \textit{Inter Miami}, which is incorrect information.

Fortunately, only a few of these issues cropped up in our tests with BlenderBot 2.0. This issue may be alleviated because BlenderBot 2.0's search engine refers to Wikipedia\footnote{\url{https://www.wikipedia.org/}}, which is one of the most active sites for adding new information and correcting with dynamic information.

\subsection{Unsafe response generation problems}
\label{sec:5_4}
We define the unsafe response problem as generating a response that raises social and ethical concerns such as profanity, racism, political and privacy issues, gender discrimination, and sexual remarks. Although this issue was rarely encountered in our tests, there is a response in CASE 4-1 of Figure \ref{fig:case_4} that demeans people of certain nationalities. Because the response was generated without the use of an internet search, we can deduce it is because the training datasets contain unsafe sentences.

\subsection{Redundant or unrelated response generation problems}
\label{5.5}
The redundant or unrelated response generation problem is repeating a previous response or generating an unrelated response. In CASE 5 in Figure \ref{fig:case_5}, The user asks about \textit{Bareilly}'s population, but BlenderBot 2.0 generates an irrelevant response (\textit{That's a good question}). It is because of the absence of persona and previous conversation history in this conversation. We can infer that BlenderBot 2.0 has some weaknesses in reflecting the multi-turn conversation.

\subsection{Tabular data problem}
The tabular data reflection problem is the inability to reflect tabular data in the response.
Figure \ref{fig:case_6} demonstrates this problem. The dialogue in left and search result in right show that an appropriate search query is generated and an appropriate internet site is discovered. When the required information is contained in the table structure, BlenderBot 2.0 does not retrieve them. Considering that Wikipedia, the main search engine of the model, it is highly likely to have fatal issues with internet searching since there are numerous tabular data.

\subsection{Numerical response problems}
The numerical response problem is defined as the inability to generate a simple number-related response. In Figure \ref{fig:case_7}, there is no direct mention of the number of children in CASE 7-1, despite three names can be obtained in CASE 7-2. Also, although the user stated that Einstein had three children in CASE 7-2, BlenderBot 2.0 response three is correct but mentions four(\textit{two sons and two daughters}). This contradiction demonstrates that BlenderBot 2.0 was unable to generate an accurate response unless the exact number for a question was specified within a site found via an internet search engine. This means that the model does not comprehend even the most basic mathematical concepts.

\subsection{URL recognition problem}
The URL recognition problem occurs when the model cannot read the information in the site when receiving a URL sequence as an input. As shown in Figure \ref{fig:case_8}, BlenderBot 2.0 recognizes URLs just as dialog sequences. The generated response and query is based on URL sequence, consequently.

In CASE 8-1 in Figure \ref{fig:case_8}, the URL of input is not retrieved because there were insufficient clues to infer what the site contained in the URL sequence. In contrast, in CASE 8-2, the URL entered by the user could not be retrieved because there were insufficient clues to infer what the site contained within the URL sequence entered by the user, so it could not generate an appropriate query. Most site URLs have a title that can infer the information of the site, but there are some sites where it is difficult to infer information from the title. Furthermore, the URL addresses do not even include titles.

\section{Discussion on how to improve BlenderBot 2.0} 
In this Section, we present approaches attempting to correct the previously mentioned errors. We propose them from the standpoints of model, data, and dialogue.

\subsection{Data and Model-centric approach-based Discussion}
\subsubsection{Reducing ambiguity in collecting data}
Crowdsourcing is used to build the MSC and WizInt dataset, which is vastly different from previous datasets. It must, however, reduce ambiguity by gathering and assembling datasets with clear standards.
 
We need to reduce the ambiguity in the MSC dataset session and clarify the criteria for crowd workers regarding when to search the internet while building the WizInt dataset. This is expected to improve the performance of long-term dialogue and internet search with new datasets built in conjunction with the previously mentioned unified standards. Furthermore, it has been confirmed that hate speech examination and filtering process will be added to alleviate related responses.

\subsubsection{Improvement for multilingual expansion}
As stated in Section \ref{sec:4_3}, dataset collection is accompanied by crowdsourcing. Due to time and financial constraints, we propose the two alternatives listed below:
 
(1) Using translation models. The best-performing model and post-editing process could be used to secure high-quality datasets. This may allow the use of previous English datasets rather than additional data collection processes.

(2) Using the model's adaptor layer. Recent research has demonstrated a cross-lingual post-training approach based on the Implicit Translation Layers (ITLs) \cite{lee2021exploring} by adapting the high resource language model into a low resource language. We can infer that this could be used to create a multilingual BlenderBot 2.0 without the need for additional crowdsourcing.

\subsubsection{Verification of Internet accuracy} 
Bing Search is used to retrieve related documents from the internet to reflect real-time information. However, why the Bing Search engine is preferred over other engines is unclear. Also, an analysis of the excellence of the engine is lacking.

Furthermore, there is no confirmation that the searched information is true, and there are no guidelines for which information to prefer the searched contexts do not match. These may be alleviated using a system that gives more weight to the most recent information among the top K documents searched.

\subsubsection{Improving for commercialization} 
The response time is likely delayed because of additional processes such as internet search and memory access to retrieve information related to dialogue contexts. To commercialize the chatbot model, the response time must be reduced to generate responses more quickly. Nonetheless, no research has been conducted on the response delay time.

It is also critical to downsize the model when using BlenderBot 2.0 for general companies or individuals. Pruning, which removes partial parameters, may help to speed up reasoning. Furthermore, knowledge distillation will deliver the same service using a smaller model trained to mimic and match the larger model.

\subsection{User-centric approach-based Discussion}
\subsubsection{Improvement in Internet document retrieval problems}
While the model generates an appropriate search query and retrieves the correct website, it does not always generate an accurate response. There are cases where the model fails to extract the correct information from the retrieved website or fails to reflect the generated information even when it does extract information correctly. 
 
The former indicates a problem with the extraction of information from the retrieved website. To address this, it is necessary to reconstruct the WizInt dataset to improve the search engine's performance by increasing the amount of data.
 
The latter indicates that, while information is extracted from the correct website, it is not properly relayed in response generation. This means that the generation model, which is an Encoder-Decoder architecture, performs poorly probably because, for the first time, the generation model used in BlenderBot 2.0 attempts to generate responses by using internet information and dialogue history. Thus, the approach that it takes is broad. This problem can be solved by increasing the amount of training data and the number of model parameters. 

\subsubsection{Improvement in search query generation problem}
The absence of a search query, even though the response should be generated using internet information, indicates that the crowd workers' knowledge is included in the crowdsourced dataset, WizInt. Furthermore, the generation of incorrect or inappropriate search queries indicates that the search query generator is underperforming. Further research on the query generator is required to solve this problem. Also, the WizInt dataset must be updated based on specific criteria.

To solve the problem of not generating a query on the first turn, we should avoid starting each turn in the data with a simple greeting.
The WizInt dataset needs to be reconfigured so that some first turns ask for information and conduct internet searches with appropriate queries.

\subsubsection{Improvement in Untrue result retrieval problems}
In general, information on the internet is accumulated rather than updated. As a result, there may be a lot of incorrect information. The best way to filter out incorrect or altered information is to use Wikipedia. Because Wikipedia is constantly modified by internet users, it is difficult to find in-accurate information, and the most recent information is also updated immediately. In fact, the WizInt dataset's majority of domains are set to the Wikipedia domain. However, because Wikipedia cannot provide all information, other domains must be used. 

As a result, when reflecting retrieved information in response generation, a filtering process is required to determine whether the information is correct or not. Searching multiple websites in a single search query and then filtering out false information based on information extracted from each website is one method to achieve highly accurate information. An alternative method is to use multiple search queries rather than just one. The model can overcome the problem of incorrect or outdated information and generate better responses using these methods.
  
\subsubsection{Improvement in duplicate responses and non-relevant response generation problem}
As described in Section \ref{5.5}, when the number of dialogue turns increases, there is an issue in which the dialogue history and persona are over-reflected in the response. This is because when creating the dataset, the dialogue history and persona are overly included in the response. To solve this issue, the dataset must be updated with specific criteria.

\subsubsection{Improvement in unsafe response generation problem}
There are two instances in which the model generates unsafe dialogue. First, an unsafe context is included in the internet-retrieved document, followed by an unsafe response in the training dataset.
In the former case, it does not occur in the inference because of the safety detector that detects the unsafe utterance. In the latter case, however, it produces unsafe utterances, which we believe is due to the training dataset. To reduce the generation of unsafe utterances, the unsafe dialogue must be removed from the dataset and the model must be designed in such a way that it does not reflect the unsafe context during encoding, such as by combining pipelines \cite{xu2020recipes}.

\subsubsection{Improvement of tabular data reflection problem}
Some Wikipedia information is in the form of tabular, as illustrated in Figure \ref{fig:case_7} c. BlenderBot 2.0 is unable to read the tabular data on the retrieved website. To reflect tabular information, a parsing algorithm or module for tabular must be added to the model encoder.  If tabular data is reflected in the encoder via pre-trained models such as TaPas \cite{herzig2020tapas} and TaBERT \cite{yin2020tabert}, the information can be reflected in the response generation.

\subsubsection{Improvement in URL recognition problem}
When given an utterance containing a URL as an input, BlenderBot 2.0 correctly creates a search query and generates a response because most URLs contain titles that can identify the website's content. However, some URLs lack text that can be used to identify the site's content. In the case of Korean articles, the title is encoded numerically using a separate algorithm and reflected in the domain. In this case, the decryption algorithm can recover the encrypted numeric domain address to the domain address, including the title. However, as illustrated in CASE 8-2, not all site domains have titles, so the search query is generated with irrelevant text. A separate module that extracts the URL from the user utterance and retrieves the extracted URL site is required to solve the URL recognition problem.

\section{Conclusions}
The open-domain chatbot, which uses a large-scale pre-trained model, naturally mimics human conversations, raising hopes that human-like AI will emerge. However, there are some constraints to having a completely human-like conversation. Although BlenderBot 2.0, which was recently released, attempted to alleviate the problems of the previous model in a variety of ways, some flaws remain. We examine the issues with BlenderBot 2.0 from the standpoints of the model, data, and dialogue, and we propose various solutions. In the future, we will fix the problems mentioned in BlenderBot 2.0 and create a Korean version of BlenderBot.

\bibliography{anthology,custom}

\begin{thebibliography}{29}
\expandafter\ifx\csname natexlab\endcsname\relax\def\natexlab#1{#1}\fi

\bibitem[{Adiwardana et~al.(2020)Adiwardana, Luong, So, Hall, Fiedel,
  Thoppilan, Yang, Kulshreshtha, Nemade, Lu et~al.}]{adiwardana2020towards}
Daniel Adiwardana, Minh-Thang Luong, David~R So, Jamie Hall, Noah Fiedel, Romal
  Thoppilan, Zi~Yang, Apoorv Kulshreshtha, Gaurav Nemade, Yifeng Lu, et~al.
  2020.
\newblock Towards a human-like open-domain chatbot.
\newblock \emph{arXiv preprint arXiv:2001.09977}.

\bibitem[{Bao et~al.(2020)Bao, He, Wang, Wu, Wang, Wu, Guo, Liu, and
  Xu}]{bao2020plato}
Siqi Bao, Huang He, Fan Wang, Hua Wu, Haifeng Wang, Wenquan Wu, Zhen Guo,
  Zhibin Liu, and Xinchao Xu. 2020.
\newblock Plato-2: Towards building an open-domain chatbot via curriculum
  learning.
\newblock \emph{arXiv preprint arXiv:2006.16779}.

\bibitem[{Baumgartner et~al.(2020)Baumgartner, Zannettou, Keegan, Squire, and
  Blackburn}]{baumgartner2020pushshift}
Jason Baumgartner, Savvas Zannettou, Brian Keegan, Megan Squire, and Jeremy
  Blackburn. 2020.
\newblock The pushshift reddit dataset.
\newblock In \emph{Proceedings of the international AAAI conference on web and
  social media}, volume~14, pages 830--839.

\bibitem[{Bender et~al.(2021)Bender, Gebru, McMillan-Major, and
  Shmitchell}]{bender2021dangers}
Emily~M Bender, Timnit Gebru, Angelina McMillan-Major, and Shmargaret
  Shmitchell. 2021.
\newblock On the dangers of stochastic parrots: Can language models be too
  big?��.
\newblock In \emph{Proceedings of the 2021 ACM Conference on Fairness,
  Accountability, and Transparency}, pages 610--623.

\bibitem[{Brown et~al.(2020)Brown, Mann, Ryder, Subbiah, Kaplan, Dhariwal,
  Neelakantan, Shyam, Sastry, Askell et~al.}]{brown2020language}
Tom~B Brown, Benjamin Mann, Nick Ryder, Melanie Subbiah, Jared Kaplan, Prafulla
  Dhariwal, Arvind Neelakantan, Pranav Shyam, Girish Sastry, Amanda Askell,
  et~al. 2020.
\newblock Language models are few-shot learners.
\newblock \emph{arXiv preprint arXiv:2005.14165}.

\bibitem[{Dinan et~al.(2018)Dinan, Roller, Shuster, Fan, Auli, and
  Weston}]{dinan2018wizard}
Emily Dinan, Stephen Roller, Kurt Shuster, Angela Fan, Michael Auli, and Jason
  Weston. 2018.
\newblock Wizard of wikipedia: Knowledge-powered conversational agents.
\newblock \emph{arXiv preprint arXiv:1811.01241}.

\bibitem[{Fong et~al.(2003)Fong, Thorpe, and Baur}]{fong2003collaboration}
Terrence Fong, Charles Thorpe, and Charles Baur. 2003.
\newblock Collaboration, dialogue, human-robot interaction.
\newblock In \emph{Robotics Research}, pages 255--266. Springer.

\bibitem[{Herzig et~al.(2020)Herzig, Nowak, M{\"u}ller, Piccinno, and
  Eisenschlos}]{herzig2020tapas}
Jonathan Herzig, Pawe{\l}~Krzysztof Nowak, Thomas M{\"u}ller, Francesco
  Piccinno, and Julian~Martin Eisenschlos. 2020.
\newblock Tapas: Weakly supervised table parsing via pre-training.
\newblock \emph{arXiv preprint arXiv:2004.02349}.

\bibitem[{Humeau et~al.(2019)Humeau, Shuster, Lachaux, and
  Weston}]{humeau2019poly}
Samuel Humeau, Kurt Shuster, Marie-Anne Lachaux, and Jason Weston. 2019.
\newblock Poly-encoders: Transformer architectures and pre-training strategies
  for fast and accurate multi-sentence scoring.
\newblock \emph{arXiv preprint arXiv:1905.01969}.

\bibitem[{Jurafsky and Martin(2019)}]{jurafsky2019speech}
Dan Jurafsky and James~H Martin. 2019.
\newblock Speech and language processing (3rd draft ed.).

\bibitem[{Karma~Choedak(2020)}]{chatbot_latency}
Karma Karma~Choedak. 2020.
\newblock The effect of chatbots response latency on users’trust.

\bibitem[{Kim et~al.(2020)Kim, Ahn, and Kim}]{kim2020sequential}
Byeongchang Kim, Jaewoo Ahn, and Gunhee Kim. 2020.
\newblock Sequential latent knowledge selection for knowledge-grounded
  dialogue.
\newblock \emph{arXiv preprint arXiv:2002.07510}.

\bibitem[{Komeili et~al.(2021)Komeili, Shuster, and
  Weston}]{komeili2021internet}
Mojtaba Komeili, Kurt Shuster, and Jason Weston. 2021.
\newblock Internet-augmented dialogue generation.
\newblock \emph{arXiv preprint arXiv:2107.07566}.

\bibitem[{Lee et~al.(2021)Lee, Yang, Whang, Park, Matteson, and
  Lim}]{lee2021exploring}
Chanhee Lee, Kisu Yang, Taesun Whang, Chanjun Park, Andrew Matteson, and
  Heuiseok Lim. 2021.
\newblock Exploring the data efficiency of cross-lingual post-training in
  pretrained language models.
\newblock \emph{Applied Sciences}, 11(5):1974.

\bibitem[{Park et~al.(2021{\natexlab{a}})Park, Eo, Moon, and
  Lim}]{park2021should}
Chanjun Park, Sugyeong Eo, Hyeonseok Moon, and Heui-Seok Lim.
  2021{\natexlab{a}}.
\newblock Should we find another model?: Improving neural machine translation
  performance with one-piece tokenization method without model modification.
\newblock In \emph{Proceedings of the 2021 Conference of the North American
  Chapter of the Association for Computational Linguistics: Human Language
  Technologies: Industry Papers}, pages 97--104.

\bibitem[{Park et~al.(2021{\natexlab{b}})Park, Park, Moon, Eo, and
  Lim}]{park2021study}
Chanjun Park, Kinam Park, Hyeonseok Moon, Sugyeong Eo, and Heuiseok Lim.
  2021{\natexlab{b}}.
\newblock A study on performance improvement considering the balance between
  corpus in neural machine translation.
\newblock \emph{Journal of the Korea Convergence Society}, 12(5):23--29.

\bibitem[{Park et~al.(2021{\natexlab{c}})Park, Seo, Lee, Lee, Moon, Eo, and
  Lim}]{park-etal-2021-bts}
Chanjun Park, Jaehyung Seo, Seolhwa Lee, Chanhee Lee, Hyeonseok Moon, Sugyeong
  Eo, and Heuiseok Lim. 2021{\natexlab{c}}.
\newblock \href {https://doi.org/10.18653/v1/2021.wat-1.10} {{BTS}: Back
  {T}ran{S}cription for speech-to-text post-processor using
  text-to-speech-to-text}.
\newblock In \emph{Proceedings of the 8th Workshop on Asian Translation
  (WAT2021)}, pages 106--116, Online. Association for Computational
  Linguistics.

\bibitem[{Radford et~al.(2019)Radford, Wu, Child, Luan, Amodei, Sutskever
  et~al.}]{radford2019language}
Alec Radford, Jeffrey Wu, Rewon Child, David Luan, Dario Amodei, Ilya
  Sutskever, et~al. 2019.
\newblock Language models are unsupervised multitask learners.
\newblock \emph{OpenAI blog}, 1(8):9.

\bibitem[{Rashkin et~al.(2018)Rashkin, Smith, Li, and
  Boureau}]{rashkin2018towards}
Hannah Rashkin, Eric~Michael Smith, Margaret Li, and Y-Lan Boureau. 2018.
\newblock Towards empathetic open-domain conversation models: A new benchmark
  and dataset.
\newblock \emph{arXiv preprint arXiv:1811.00207}.

\bibitem[{Roller et~al.(2020)Roller, Dinan, Goyal, Ju, Williamson, Liu, Xu,
  Ott, Shuster, Smith et~al.}]{roller2020recipes}
Stephen Roller, Emily Dinan, Naman Goyal, Da~Ju, Mary Williamson, Yinhan Liu,
  Jing Xu, Myle Ott, Kurt Shuster, Eric~M Smith, et~al. 2020.
\newblock Recipes for building an open-domain chatbot.
\newblock \emph{arXiv preprint arXiv:2004.13637}.

\bibitem[{Smith et~al.(2020)Smith, Williamson, Shuster, Weston, and
  Boureau}]{smith2020can}
Eric~Michael Smith, Mary Williamson, Kurt Shuster, Jason Weston, and Y-Lan
  Boureau. 2020.
\newblock Can you put it all together: Evaluating conversational agents'
  ability to blend skills.
\newblock \emph{arXiv preprint arXiv:2004.08449}.

\bibitem[{Song et~al.(2019)Song, Zhang, Cui, Wang, and
  Liu}]{song2019exploiting}
Haoyu Song, Wei-Nan Zhang, Yiming Cui, Dong Wang, and Ting Liu. 2019.
\newblock Exploiting persona information for diverse generation of
  conversational responses.
\newblock \emph{arXiv preprint arXiv:1905.12188}.

\bibitem[{Weizenbaum(1966)}]{weizenbaum1966eliza}
Joseph Weizenbaum. 1966.
\newblock Eliza—a computer program for the study of natural language
  communication between man and machine.
\newblock \emph{Communications of the ACM}, 9(1):36--45.

\bibitem[{Xu et~al.(2020)Xu, Ju, Li, Boureau, Weston, and
  Dinan}]{xu2020recipes}
Jing Xu, Da~Ju, Margaret Li, Y-Lan Boureau, Jason Weston, and Emily Dinan.
  2020.
\newblock Recipes for safety in open-domain chatbots.
\newblock \emph{arXiv preprint arXiv:2010.07079}.

\bibitem[{Xu et~al.(2021)Xu, Szlam, and Weston}]{xu2021beyond}
Jing Xu, Arthur Szlam, and Jason Weston. 2021.
\newblock Beyond goldfish memory: Long-term open-domain conversation.
\newblock \emph{arXiv preprint arXiv:2107.07567}.

\bibitem[{Yin et~al.(2020)Yin, Neubig, Yih, and Riedel}]{yin2020tabert}
Pengcheng Yin, Graham Neubig, Wen-tau Yih, and Sebastian Riedel. 2020.
\newblock Tabert: Pretraining for joint understanding of textual and tabular
  data.
\newblock \emph{arXiv preprint arXiv:2005.08314}.

\bibitem[{Zhang et~al.(2018)Zhang, Dinan, Urbanek, Szlam, Kiela, and
  Weston}]{zhang2018personalizing}
Saizheng Zhang, Emily Dinan, Jack Urbanek, Arthur Szlam, Douwe Kiela, and Jason
  Weston. 2018.
\newblock Personalizing dialogue agents: I have a dog, do you have pets too?
\newblock \emph{arXiv preprint arXiv:1801.07243}.

\bibitem[{Zhong et~al.(2020)Zhong, Zhang, Wang, Liu, and
  Miao}]{zhong2020towards}
Peixiang Zhong, Chen Zhang, Hao Wang, Yong Liu, and Chunyan Miao. 2020.
\newblock Towards persona-based empathetic conversational models.
\newblock \emph{arXiv preprint arXiv:2004.12316}.

\bibitem[{Zhou et~al.(2020)Zhou, Gao, Li, and Shum}]{zhou2020design}
Li~Zhou, Jianfeng Gao, Di~Li, and Heung-Yeung Shum. 2020.
\newblock The design and implementation of xiaoice, an empathetic social
  chatbot.
\newblock \emph{Computational Linguistics}, 46(1):53--93.

\end{thebibliography}
\bibliographystyle{acl_natbib}

\end{document}